\documentclass{article}
\usepackage[utf8]{inputenc}

\usepackage[]{geometry}

\usepackage[fleqn]{amsmath}
\usepackage{amsthm}
\usepackage{amsfonts}
\usepackage{amssymb}

\usepackage{titlesec}
\usepackage{color}
\usepackage[]{graphicx}
\usepackage{caption}
\usepackage{subcaption}
\usepackage{wrapfig, framed}
\usepackage[many]{tcolorbox}

\usepackage{graphicx}
\usepackage[square,sort,comma,numbers]{natbib}

\usepackage{floatpag}
\usepackage{fancyhdr}

\title{What does it mean to understand a neural network?}

\author{Timothy P. Lillicrap \& Konrad P. Kording}
\date{July 2019}

\begin{document}

\maketitle

\begin{abstract}
We can define a neural network that can learn to recognize objects in less than 100 lines of code. However, after training, it is characterized by millions of weights that contain the knowledge about many object types across visual scenes. Such networks are thus dramatically easier to understand in terms of the code that makes them than the resulting properties, such as tuning or connections. In analogy, we conjecture that rules for development and learning in brains may be far easier to understand than their resulting properties. The analogy suggests that neuroscience would benefit from a focus on learning and development.
\end{abstract}

\section*{Introduction}
When we build networks that solve image recognition at human-like performance~\cite{krizhevsky2012imagenet} or are strong at playing the game of Go~\cite{silver2017mastering} we ended up using a few computer screens worth of high-level computer code. We undeniably understand these lines of code, and we teach how these systems works to students in our deep learning courses who also obtain a meaningful understanding. After training we have the full set of weights and elementary operations. We can also compute and inspect any aspect of the representations formed by the trained network.  In this sense we can have a complete description of the network and its computations. And yet, neither we, nor anyone we know feels that they grasp how processing in these networks truly works.  Said another way, besides gesturing to a network's weights and elementary operations, we cannot say how it classifies an image as a cat or a dog, or how it chooses one Go move over another. For neural networks there is no doubt that the understanding that we can currently have about their properties after learning is massively more shallow than the understanding that we have about the code used to train it: the rules for its development and learning.

One day, we may develop ways of compactly describing how neural networks work after training.  There may be an intermediate language in which we could meaningfully describe how these systems work. A useful intermediate language would allow us to write high-level computer code to directly construct powerful neural networks for image recognition or playing Go without resorting to iterative construction of the network via a learning procedure.  However, we can not currently offer such an intermediate description for our artificial networks, which we have arbitrarily better access to than brains. As such, even if meaningful descriptions  exist, for the moment we best understand neural networks in terms of their mechanisms for development and learning. 

\begin{figure}
    \centering
    \includegraphics[width=0.9\textwidth]{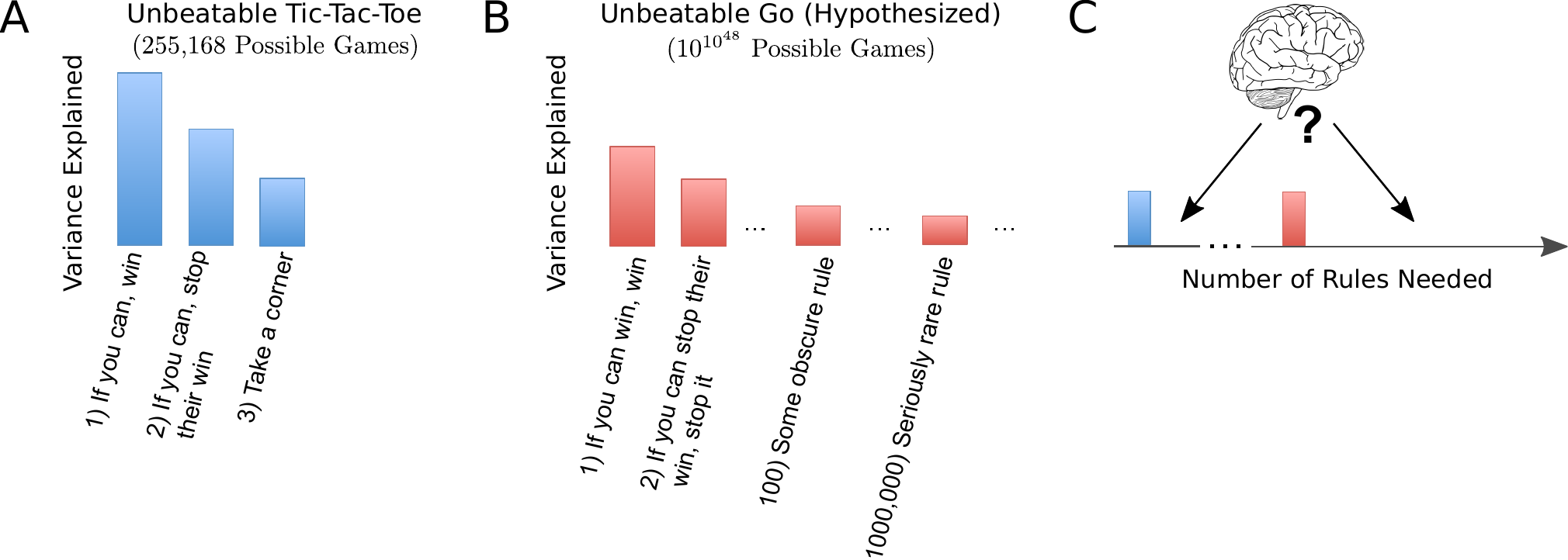}
    \caption{\textbf{The notion of compressability.} (A) Unbeatable performance at tic-tac-toe can be obtained with just three rules. These rules can be ordered by their importance  (B) Unbeatable performance at the game of Go needs many rules. So many, in fact, that we can not know that number. Importantly, the rules are heavy tailed. If we have any small number of rules, the remaining rules will generally still carry a lot information about how to further improve on the task. Humans have long tried to distill the game of Go into instructional books and have, to a large extent, been unsuccessfully at doing so. (C) The question is where on this axis the brain is situated. The brain might be easier than Go to compactly describe. Or the brain may be much harder than Go, affording no compact description under any realistic assumption.  }
    \label{fig:1}
\end{figure}

In this paper, we will discuss issues of complexity as applied to neural networks and other artificial systems. These systems are fully observed, we have a total description of all the involved functions, and nonetheless we have trouble producing a meaningful ``understanding''. We will use this consideration as a backdrop to ask what we mean when we talk about understanding neural computation. We will argue that the brain's generative process is not that unlike that of neural networks, since it obtains information from a world which it stores as a distributed pattern of weight changes, a pattern that is remarkably hard to wrap one's head around. We will conclude by suggesting that, at least for the moment, neuroscience should focus on understanding development and learning.


\section*{Can an artificial system that is as good as humans at a broad range of tasks be compactly expressed?}

The learning rules in artificial systems are compactly expressed, in the sense that they can be written in a few pages of high-level (e.g. Pytorch) code. And at least to us practitioners of machine learning it feels like having such a compact description is necessary for meaningful understanding. Having such a compact description offers us leverage: we can produce variants of the same kind of ideas and use them to solve a broad range of useful and interesting machine learning problems. Maybe humans or other intelligent beings can figure out ways of meaningfully arguing about non-compact systems, but, at least so far, compactness is necessary for what we would call a meaningful understanding. 

Central to our arguments is the idea of compressibility. We will introduce it with with a simple scenario: the game of tic-tac-toe. While there are 255,168 potential games, we can have a policy of just three rules (see Figure 1), that allows unbeatable tic-tac-toe performance. In the game of Go, in which computers have recently started beating world champions~\cite{silver2016mastering, silver2017mastering}, there is no known compact way of expressing the instantaneous valuation of a position. This is at least in part due to the fact that there can be exponentially many games of Go ( $>10^{10^{48}}$ games) although the set of reasonably probable games is probably much lower, say 800 moves, 100 possibilities each ($10^{1600}$ games). In such a domain, knowledge that lowers the number of allowable games by a factor of two (hypothesis testing) will be essentially useless (we will still have $10^{10^{48-\log_{10}(2)}}$ games left). Neural networks that capture effective policies and value functions for Go reflect the complexity of the game.  Systems that solve a broad range of interesting problems at human levels should be less compressible than specialized systems that play Go~\cite{silver2017mastering,silver2016mastering} or neural networks that recognize visual categories~\cite{krizhevsky2012imagenet}. After all, humans can do both.  In complex domains, approximate systems that are more compressed tend to have poor performance and so do not truly represent a way of avoiding the problem. We may thus expect broadly applicable intelligent systems to be hard (or impossible) to compress into a small size.


Computer scientists have worked extensively on the problem of compressing large neural networks into simpler and more describable systems. With the rise in popularity of distillation techniques~\cite{hinton2015distilling, han2015deep, howard2017mobilenets, iandola2016squeezenet, schmidt2009distilling, zhou2018non} and renewed interest in the field of computational complexity we are starting to understand more about compressability. From distillation techniques we know that networks trained on ImageNet, a popular 2012 machine learning benchmark that requires the classification of natural images, cannot readily be compressed to fewer than about 100k free parameters~\cite{hinton2015distilling,wu2016binarized, li2018measuring} (though see~\cite{zhou2018non}). We want to emphasize that these compressions, even of the ImageNet case, are not in any meaningful sense human understandable. Even the famous and somewhat trivial networks that solve the MNIST character recognition problem can not readily be compressed into a format that humans find readable~\cite{frosst2017distilling}. Moreover, human performance includes knowledge of roughly 30,000 categories of objects~\cite{biederman1987recognition} - and arguably they, on average, know a lot of information about each of them, possibly megabytes thereof, and certainly more than a few bytes. Humans also appear to know more than a megabyte worth of information about their own language~\cite{mollica2019humans}. Human-like performance seems exceptionally hard to compress into a compact representation.

How are neural network scientists trying to understand their networks? They primarily look for phenomena relevant for optimization, such as vanishing and exploding gradients, or exploding global dynamics during the process of generating a system. Perhaps most common in practice is looking at histograms of the unit activations and derivatives~\cite{bengio1994learning,glorot2010understanding,pascanu2013difficulty} to see whether learning is halting or being slowed for a trivial reason -- offering clues as to how to adjust learning rates or other hyperparameters.  However, such analyses are used almost exclusively to diagnose problems in the engineering process that arise during the construction of the system. They do not provide a path to a meaningful understanding of the computations done by the systems they build.


Ongoing work in the deep learning field aims at more intuitive understanding of the computations in neural networks. Scientists study the sensitivity of outputs to changes in the system to ask what matters~\cite{szegedy2013intriguing,gatys2016image,yosinski2015understanding}. They ask which stimuli can fool a system~\cite{szegedy2013intriguing,goodfellow2014explaining,athalye2017synthesizing}. They visualize elements of a network~\cite{zeiler2014visualizing,yosinski2015understanding}. They analyze what happens if a system is perturbed, e.g. by removing units~\cite{lecun1990optimal, olden2002illuminating, han2015deep}.  However, no one who is familiar with these approaches would say that they offer a good understanding of models like AlexNet, AlphaGo, or GPT2~\cite{krizhevsky2012imagenet, silver2017mastering,radford2019language}. These approaches remain far from offering practitioners enough understanding to be useful to improve networks for task performance.

Returning to the question of compressibility, we may ask why we should expect neural networks that exhibit human-like performance to be hard to compress? The world in which humans live is undeniably complicated; much of our world is built of peculiarities which are obviously incompressible. Let us choose balloons as an example. Adult humans know a lot about balloons. For example, they are made of rubber or aluminum foil, often held by kids or clowns, are red in a famous song, etc. And yet, it seems unlikely that we are born with knowledge of balloons. We have similar incompressible knowledge about any of the many categories of objects people know about. Estimates of how much humans know vary widely but suggest that compressibility is limited. The important result is that {\em the bulk} of the information we need to specify a working model comes from the world, which is complicated and (arguably) not compressible. There are domains where the world is simple and then we should be good at building models (e.g. for the vestibuloocular reflex~\cite{robinson1976adaptive}) and there are domains where the world is truly complicated (e.g. image recognition) and it is presumably hopeless to communicate a working model of such behaviors.

Given just the architecture and weights of an already functioning system is not very useful except with respect on the exact task that it already solves.  (Imagine giving an ImageNet trained network to someone who had no idea how the learning process worked!).  Knowing the learning rules, architectures, and loss functions used to train such a system are far more useful for future alterations and interventions one might want to make.

\section*{Analogy to neuroscience}
Humans are not neural networks. And yet, the brain has ubiquitous plasticity~\cite{kandel2000principles}. Specifically, we know that plasticity allows changes in the brain to enable good performance across new tasks~\cite{fetz1969operant}. As such, it is hard to see how the arguments we made above about artificial neural networks would not carry over to the human brain. To the level that this insight holds, this may suggest that understanding representations, which may be viewed like an artificial neural network after training, may be outstandingly hard. We concluded above that for artificial neural networks a focus on objectives, learning rules, and architectures is most promising. We do not see why the same argument does not carry over to neuroscience: itt may currently be prohibitively hard to understand connection strengths and representations, even if architectures, learning, and development can be meaningfully characterized and communicated.

\section*{Where to now?}

However, while we may not be very hopeful about communicating a working model of the dynamics, we can still hope to communicate how biology, nature, sets up the brain (including its anatomy and learning rules, see Figure 2). After all, the complexity from our DNA is upper limited by $6 \cdot 10^{9}$ bits of DNA. And that information is hugely redundant, largely non-coding, and only effects change through some 20k genes, most of which undoubtedly do not code information for brains. That being said, we can not know the amount of information effectively communicated by non-coding parts of the DNA. With the increased availability of genetic tools we might be able to get closer to such principles. The basic biochemistry of cells, is somewhat bounded in complexity -- there are only so many types of neurotransmitters or channels that appear relevant. The developmental processes setting up our anatomy as well as cell types (or computational primitives) promise to produce the backbone for our learning. We may hope that we can understand them. While learned information is unlikely to be compressible, we may at least hope that biological information is, e.g. learning rules and anatomy.

The breakdown into parameters (nurture) and principles (nature) is precisely the breakdown that the AlphaGo and AlexNet papers use in practice: they offer pointers to tasks and related data sources (from which the vast majority of the incompressible \& non-communicable parameters will be extracted), and then document the architectures (Convnets / ResNets), loss functions (L2 / cross entropy), learning rule (SGD / Adam), and optimization (MCTS) approaches used.  These last four pieces are all easily expressed in small human understandable equations, and formal recipes. We can not communicate the data sources in any straightforward sense - while we could read the contents of the images and their labels to another person, they would not meaningfully understand their totality. And yet, we do understand the generation of the neural networks.  Not only do these recipes offer replicable means of producing working systems, they also offer the kind of understanding that affords ways of pursuing useful and interesting interventions and predictions.  


There is a widespread belief in neuroscience that there can be a meaningful mid-level model of the dynamics of the brain that both can be communicated and work for complex tasks. The idea is in analogy to physics: In quantum mechanics, the state of e.g.~a gas can not be meaningfully communicated since every gas atom may exist in a high dimensional space and there are many atoms. But in statistical physics, the state of the gas may be compactly described in terms of a few variables like temperature, pressure, etc.  And these mid-level descriptions can afford high accuracy predictions and control. From this analogy it is sometimes argued that a compact mid-level description should exist for neuroscience. And, indeed, a lossy model of how e.g. eye movement is computed in humans might enable us to better predict movements for some practical purpose. Such mid-level lossy descriptions of the computations carried out by brains may thus be interesting. However, for the more complicated problems we discuss in this paper, the analogy quickly breaks down: in the gas case, all atoms are the same, are exchangeable, and have short memory while in brains each cell may be unique and have a memory that effectively goes back to the birth of the animal.  Moreover, the argument we made here suggests that such a compact mid-level model of computation can not have the property of actually working in the domains of brain performance where the environment can not be compactly communicated. Thus the analogy to physics may be misleading in the context of neuroscience. 

\section*{Outlook}




\begin{figure}[h!]
    \centering
    \includegraphics[width=0.9\textwidth]{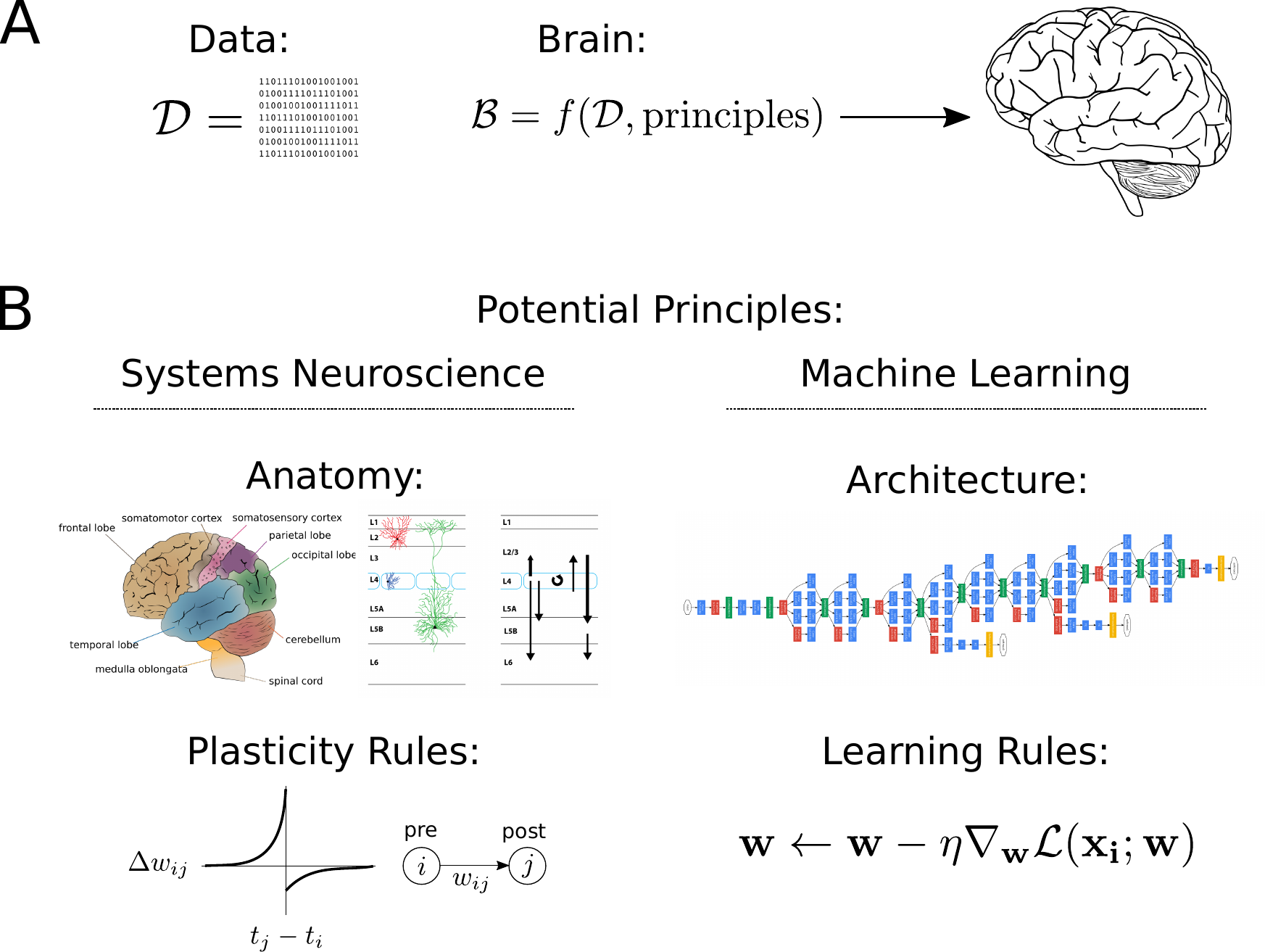}
    \caption{\textbf{Dividing theories of brain functions into principles and data.} 
    (A) If the brain is not compressable, we can at best divide our description of it into a set of compact principles and a set of non compressable data. We may then hope that the way the brain converts the data into computation may be understandable. (B) A very natural division is to ask for an understanding of anatomy and plasticity rules, which jointly may allow understanding how (uncompressable) stimuli convert into computation. This could be in analogy to machine learning where architectures along with learning rules and loss functions give rise to the relevant computation.}
    \label{fig:2}
\end{figure}

History is full of seemingly impossible things that ended up being possible~\cite{dickinson2001solving}. It may thus eventually be possible to meaningfully understand a network trained on ImageNet. Importantly, there can be no doubt that doing so is arbitrarily easier than understanding a human brain - after all, for a network trained on ImageNet we have all the information and can run any experiment we might possibly  want to run. We thus may want to try and start trying to understand such networks, in tight analogy with microprocessors~\cite{jonas2017could}. Despite a thriving community aiming at understanding trained neural networks \cite[e.g.][]{han2015deep,farrell2019dynamic} there are worries about general feasibility~\cite{ramaswamy2019algorithmic}.  It is possible that real brains have aspects, e.g. strong modularity, approximate linearity, or high noise that make them easier to understand than artificial neural networks, in which case we should be precise about which factors make the task doable.  Alternatively, it may be impossible to meaningfully understand a network after training on a complex tasks and we may want to start looking for ways of proving that. Or maybe humans can never understand such a network but, potential super-human entities, e.g. those built by future AI technologies, may be able to understand it. In any of these cases, neuroscience should be informed by this consideration and a lot of approaches currently deployed to understand brains may be rather transparently unable to deliver the results neuroscientists are looking for. 

The popular approaches used to study brains can readily be focused on studying the nature component as opposed to nurture component. Behaviorally, we may ask how learning changes the choice of actions. When we study representations, we can ask which loss function, architecture, and learning dynamics would have given rise to the measured representations. Alternatively, we may ask about the representation of learning signals. When we study connectomics, we can focus on the large scale anatomy and circuit motifs~\cite{rubinov2010complex,douglas1991functional}, which likely are genetically specified, instead of which exact neuron connects to which other exact neuron, which is likely to be learned. Perturbation approaches such as optogenetics can be focused on perturbing the signals that matter for learning. Even within molecular research we may focus on the cascades that set up the computation, e.g. synaptic plasticity, versus those involved in the computation per se. Instead of asking {\em how the brain works} we should, arguably, ask {\em how it learns to work}.


\section{Acknowledgements}
We are thankful to many people for discussion and inspiration. Of particular importance were John Krakauer, Neil Rabinowitz, the MILA group, and Adam Marblestone. We want to thank NIH (R01MH103910) for funding.

\bibliographystyle{plain}
\bibliography{references}

\end{document}